\documentclass[11pt]{article}

\usepackage[final]{acl}

\usepackage{times}
\usepackage{latexsym}

\usepackage[T1]{fontenc}

\usepackage[utf8]{inputenc}

\usepackage{microtype}

\usepackage{inconsolata}

\usepackage{graphicx}

\usepackage{amsmath}
\usepackage{amssymb}
\usepackage{booktabs}
\usepackage{makecell}
\usepackage{tikz}
\usepackage{array}
\usepackage{colortbl}
\usepackage{xcolor}
\usepackage{multirow}
\usepackage{tikz}
\usetikzlibrary{shapes.geometric, arrows.meta, positioning, fit, backgrounds, calc, shadows, shapes.multipart}

\usepackage[table]{xcolor}
\definecolor{lightblue}{RGB}{173,216,230}
\definecolor{headercolor}{RGB}{70,130,180}
\usepackage{graphicx}
\usepackage{adjustbox}

\title{CLUE: A Contrastive Learning Framework for Unsupervised Event Type Induction and Annotation}

\author{
Pratik Maitra, Lauren Kwon, Qi Li \\
Iowa State University, Ames, IA, USA \\
\texttt{\{pmaitra, isoo0301, qli\}@iastate.edu} \\
}

\begin{document}
\maketitle

\begin{abstract}
Event ontology is the foundational framework for defining events in a dataset. Event type induction generally involves domain experts who define guidelines for creating event types and their names based on the dataset's domain and scope. However, this process is resource-intensive in terms of annotation costs and is prone to human biases. To alleviate the need for manual event ontology creation, we propose the \textbf{CLUE} framework for event type induction, naming, and annotation. Our main contributions in this work are: 1) designing a novel joint embedding representation for expressing both the verb-centric view and the semantic view of events; 2) event type induction using contrastive loss guided clustering; 3) event naming and refinement. Experiments demonstrate the effectiveness of our method\footnote{Our code is available at \url{https://github.com/PratikMaitra/CLUE}}.
\end{abstract}



\section{Introduction}
An event is a specific action that involves certain participants or objects, and event extraction is a key part of information extraction~\citep{BachAlgebra, sep-events, sundheim-1992-overview, doddington-etal-2004-automatic}. To construct an event extraction data set, the first step is to define the event ontology for the target corpus, which consists of the set of relevant event types and their corresponding names~\citep{brown-etal-2017-rich}. The event type induction task is the problem of discovering the set of salient events in a corpus~\citep{shen-etal-2021-corpus}.
\begin{figure}[t]
\centering
\renewcommand{\arraystretch}{1.25}
\setlength{\tabcolsep}{2pt} 
\small  

\begin{tabular}{|>{\centering\arraybackslash}p{0.5cm}|p{4.4cm}|p{1.9cm}|} 
\hline
\rowcolor{headercolor}
\textbf{\textcolor{white}{Sid}} &
\textbf{\textcolor{white}{Sentence}} &
\textbf{\textcolor{white}{P--O}} \\
\hline
\rowcolor{lightblue}
S1 & Bob set the table for dinner. & set - table \\
\hline
\rowcolor{lightblue}
S2 & Bob set a new record in tennis. & set - record \\
\hline
\rowcolor{lightblue}
S3 & Bob set the new record in the vinyl player. & set - record \\
\hline
\rowcolor{lightblue}
S4 & Bob arranged the table for supper. & arrange - table \\
\hline
\rowcolor{lightblue}
S5 & Bob achieved a new milestone in chess. & achieve - milestone \\
\hline
\end{tabular}
\caption{Polysemy of the verb \emph{set}. Sentences S1 and S4 describe the same underlying event, as do S2 and S5. Meanwhile, S2 and S3 share a similar predicate--object pair but differ in meaning.}
\label{fig:fig1}
\end{figure}

Pre-transformer era approaches to the same or related tasks include clustering-based methods relying on event templates~\citep{chambers-jurafsky-2011-template}, generative methods that use topic distributions~\citep{nguyen-etal-2015-generative}, and semantic and lexical methods that rely on trigger words and event descriptions to label event types~\citep{bronstein-etal-2015-seed}. 

Recent work in the unsupervised event type induction task has leveraged transformer-based language models to cluster predicate-object (P-O) pairs extracted from the sentences of the corpus~\citep{shen-etal-2021-corpus, xu-etal-2024-ceo}. A key challenge in this task is the polysemous nature of predicate verbs. For example, the verb "set" in sentences S2 and S3 of Figure~\ref{fig:fig1} illustrates how identical P-O pairs can express different meanings and relate to different underlying events. Sentence S2 refers to the event of \textit{"setting a record in sports"}, while sentence S3 refers to the event of \textit{"setting a music record in the vinyl player"}. Prior work used external resources, such as verb dictionaries or summary datasets, to handle polysemy and disambiguate verb senses, but these resources are created through expensive manual annotation and are often limited to specific domains. Moreover, these methods only induce the event types from the corpus, but do not annotate sentences with the discovered event types.

In this work, we propose \textbf{CLUE}, an unsupervised framework that annotates a text corpus and induces major event types while handling polysemous verbs at low annotation cost. Instead of querying an LLM for each sentence annotation, CLUE first clusters sentences in the corpus in order to choose the representative ones for LLM annotation. With the LLM provided synthetic supervision on event type names and descriptions, CLUE further refines the event types and annotations by merging redundant clusters based on the semantic similarity of event type names and descriptions. 

To address the polysemy issue of verbs, CLUE preserves the original sentence with the P-O pairs to retain semantic context and disambiguate verb senses. To increase the diversity of sparsely represented predicate verbs, we also adopt a mask-token-prediction augmentation method on predicate verbs. Then the original and augmented instances are jointly used to train a better P-O pair representation using contrastive learning. These fine-tuned P-O pair representations can better yield semantically coherent and distinct event clusters. 

 We summarize our main contributions in the paper as follows:
\begin{enumerate}
    \item A joint framework that can induce event types in an unsupervised manner and annotate sentences with these event types.
    \item A domain-agnostic contrastive clustering method that uses augmentation to handle polysemy and represent long-tail predicate verbs.
    \item An event naming and refinement approach that generates the major event types and enables corpus-level annotation.
\end{enumerate}

\section{Related Work}
Event extraction and detecting event mentions in a dataset have been a top-level focus of NLP research. Several span-based methods~\citep{wadden-etal-2019-entity}, clustering based methods~\citep{rt_ed}, supervised classification methods~\citep{hsu-etal-2022-degree, li-etal-2023-glen, peng-etal-2023-devil}, and generative methods using LLMs~\citep{Chen_Qin_Jiang_Choi_2024, dong-etal-2025-scievent, parekh-etal-2025-snare} have been investigated in previous works in this area. Research on event datasets and dataset pre-processing techniques has also been explored in prior works~\citep{lu-etal-2023-event, huang-etal-2024-textee, wei-etal-2025-legalcore}.

\noindent\textbf{Event ontology creation and alignment:} Event ontology creation has traditionally been done following manually created annotation guideline documents, as seen in the ACE~\citep{doddington-etal-2004-automatic} and GENIA~\citep{kim-etal-2013-genia} datasets. Event ontologies differ between domains, and previous work has examined the creation of domain-specific ontologies~\citep{bonial-etal-2016-multimodal, brown-etal-2017-rich, news_event, segers-etal-2017-circumstantial, guo-etal-2023-eventoa}.

\noindent\textbf{Weak and semi-supervised event type induction:} To mitigate the time and resource constraints of human-guided event ontology construction, some researchers have explored weak and semi-supervised event type induction, which utilizes both seen and unseen event types \citep{lin-etal-2020-joint, huang-ji-2020-semi, li-etal-2022-open,  cao-etal-2023-event}.

\noindent\textbf{Unsupervised event type induction:} The unsupervised event type induction task induces event clusters without relying on known/seen event type labels and this is first explored in EtypeClus, which addresses the top-level event types or the flat event ontology~\citep{shen-etal-2021-corpus}. However, EtypeClus does not name the event clusters. In a more recent work, the authors used distant supervision from summary datasets to obtain hierarchical event types, which are subsequently named~\citep{ xu-etal-2024-ceo}.

\section{Task Formulation and Definitions}

The given corpus is designated as $\mathcal{C}$, which has $J$ sentences such that $\mathcal{C} = \{ S_1, \ldots, S_J \}$. 


\textbf{Event type induction} is formalized as the task of discovering event types ($E^T$) in a corpus ($\mathcal{C}$). In the unsupervised setting, the induction task does not have access to known event types.


\textbf{Event type naming} is defined as the task of assigning an interpretable and human-readable event type name to an event, $E^{N}_i = f_{\text{name}}(E^{T}_i)$.


\textbf{Event type annotation} is the task of assigning the most appropriate event type label(s) to a sentence. 

In this paper, we aim to conduct all three tasks using one framework. 

\begin{figure*}[t]
  \centering
  \includegraphics[width=\linewidth]{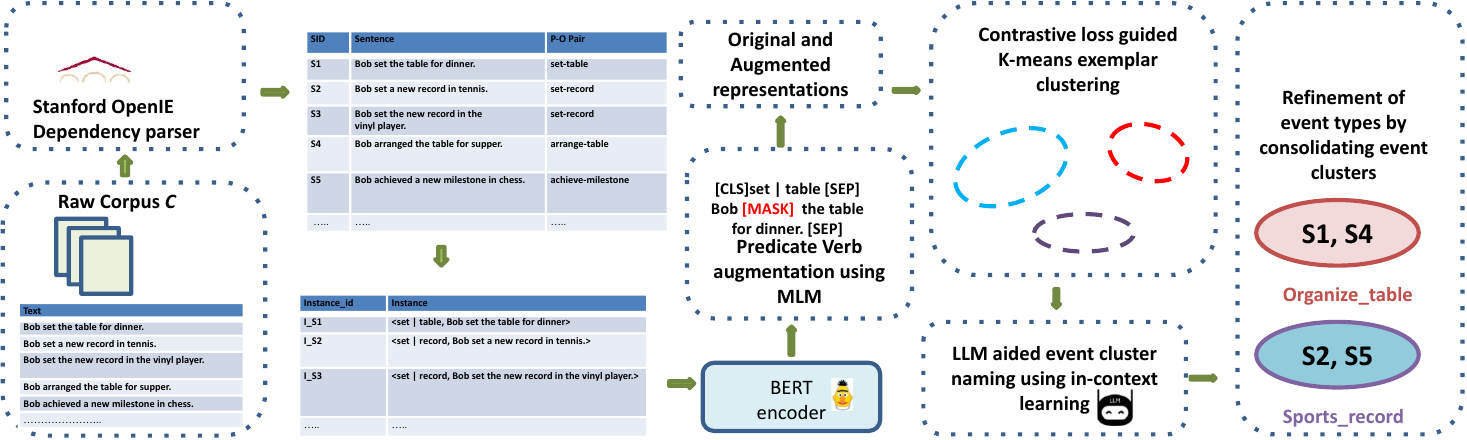}
  \caption{Overview of CLUE. The event type induction consists of 1) P-O pair extraction by a parser, 2) Joint instance creation, 3) BERT encoding of the joint instances, 4) Augmentation of the predicate verbs using MLM, 5) applying contrastive learning guided, 6) K-Means exemplar clustering. The event induction step is followed by 7) event type naming using an LLM and 8) event type refinement.}
  \label{fig:fig2}
\end{figure*}

\section{Methodology}
An overview of the CLUE framework is shown in Figure~\ref{fig:fig2}. The event type induction consists of 1) extracting predicate verb object lemma templates, 2) creating joint instances of Predicate Verb Object Lemma templates, 3) BERT encoding of the joint instances, 4) augmentation of the predicate verbs, 5) fine-tuning the BERT representation using contrastive learning, and 6) applying K-Means exemplar clustering for inducing events. The event induction step is followed by 7) event type naming and 8) event type refinement, in which the initial clusters are merged and consolidated based on the semantic similarity of their names.

\subsection{Extracting Predicate Verb Object Lemma Templates}
In the unsupervised event induction task, the predicate verb and object-head lemma pair carry key significance for event type induction. The predicate verb of a sentence represents the action that is being performed on the object and is often the "trigger" that signifies the event type. We extract the predicate verb and object head lemma from each sentence using the Stanford OpenIE~\citep{angeli-etal-2015-leveraging} library and a dependency parser. We also filter out common and trivial verbs during the extraction stage.

\subsection{Joint Instance of Predicate-Object Pair and Original Sentence}
Corpus-observed predicate verbs alone provide limited lexical coverage for the event cluster induction task. The key challenge is to capture all the diverse instances of a trigger word related to a specific event type. Prior work used external resources such as summarization datasets and verb banks to derive salient verb lemmas that are semantically similar~\citep{xu-etal-2024-ceo}.

CLUE instead uses the original sentence from which the P-O pair is extracted to provide context. The sentence contains key contextual information that determines the semantic nature of the predicate verb. The instance for our model is thus designed as a joint embedding, where the predicate-verb lemma and the object-head lemma are separated by a $|$ and attached to the original sentence. The joint instance is used for both augmenting predicate verbs and inducing event clusters.

The instance is formally defined as follows:
\begin{equation}
\mathbf{inst}
=
\langle p \mid o, \; S \rangle
\end{equation}

\subsection{BERT encoding of the joint instances}
Our instance is encoded using a PLM representation. The \textsc{BERT}\textsubscript{base}\ model is chosen to encode our instances \citep{devlin-etal-2019-bert}. The \textsc{BERT}\textsubscript{base}\ model has seen widespread use in NLP tasks and is the default choice for PLM representation. Our instances, consisting of the predicate verb lemma $p$, the object head lemma $o$, and the original sentence $S$ that contained the predicate verb and the object, are represented as the embedding $\mathbf{h}_{\text{inst}}$.
\begin{equation}
\mathbf{h}_{\text{inst}}
=
\text{BERT}_{\text{base}}\big(
\texttt{[CLS]} \ p \mid o \ \texttt{[SEP]} \ S \ \texttt{[SEP]}
\big)
\end{equation}

\subsection{Augmentation of Predicate Verbs}
An event type can be associated with multiple diverse event trigger words or predicate verbs. For instance, there may be synonymous verbs like "hit" and "strike" that are related to the same event type, but only one of them may be present in our corpus. Certain predicate verbs related to a specific event type might have a sparse representation in the corpus. 

This motivates us to augment the list of initial predicate verb lemmas extracted from the sentences. Context-free augmentation can suffer from polysemy. As the joint instance, composed of the P-O pair and the sentence, holds key semantic and contextual information, the augmentation process addresses polysemy by producing a diverse list of semantically similar predicate verbs. It is also important to note that we do not augment or modify the object head, as it is often a named entity or a noun phrase. The object head lemma does not possess the same semantic weight as the predicate verb.

The augmentation is achieved by masking the token corresponding to the predicate verb in the sentence contained in our joint instance embedding. The original and augmented instances form positive pairs for representation learning. 

For a sentence $S$ in our corpus, our augmented embedding is as follows:
\begin{equation}
\begin{aligned}
\mathbf{h}_{\text{aug}} 
&= \text{BERT}_{\text{base}}\big(
\texttt{[CLS]}\; p \mid o\; \texttt{[SEP]} \\
&\quad S_{\text{masked}}\; \texttt{[SEP]}
\big).
\end{aligned}
\end{equation}
where in $S_{\text{masked}}$, \texttt{[MASK]} replaces $p$, which spans positions $(i,j)$ in $S$.
\begin{equation}
S_{\text{masked}} = S[1{:}i] \oplus \texttt{[MASK]} \oplus S[j{:}n]
\end{equation}

\subsection{Contrastive Learning with Margin Loss}
Another challenge of the event induction task is obtaining clusters that contain text which is semantically coherent within the same cluster but distinct in semantic content from the text in other clusters. This can be achieved by using a learning paradigm that leverages the semantic and contextual views of our instance representation. 

The multi-view learning paradigm best suited for our framework is contrastive learning. Contrastive learning is a representation learning framework that has gained popularity in recent works due to its ability to pull together similar representations and push away differing representations in the embedding space~\citep{Oord2018RepresentationLW, DBLP:conf/iclr/0001ZXH21,DBLP:journals/corr/abs-1911-05722}.

Traditionally, contrastive learning has utilized the InfoNCE loss function, but contrastive margin loss has been found to be more robust to noise for similar semantic clustering tasks, as seen in previous work~\citep{wang-etal-2023-improving-unsupervised}. The contrastive margin loss function is utilized over other loss functions in clustering similar event-type sentences. The contrastive margin loss is defined as follows:
\begin{equation}
\begin{aligned}
L_m=\frac{1}{M}\sum_{i=1}^{M}
\max\Bigl\{&
\mathrm{dist}(a_i,p_i)-\mathrm{dist}(a_i,n_i) \\
&+\gamma,\;0
\Bigr\}.
\end{aligned}
\label{eq:loss}
\end{equation}
Here, $a_i$, $p_i$, and $n_i$ denote the representations of the anchor,
positive, and negative instances, respectively, where each instance
corresponds to a pair of joint predicate verb–object lemmas concatenated 
with its sentence context. The function $\mathrm{dist}(\cdot)$ measures
the cosine distance between the representations of instances, and $\gamma$
denotes the margin hyperparameter.

\subsection{K-Means Exemplar Clustering for Event Type Induction}
The event-type induction is to be performed by an algorithm that can efficiently group textual data into event clusters in the unsupervised setting. As discussed in the previous section, the grouping would be aided by a view-based contrastive learning principle. 

Given the unsupervised nature of the event induction task and our joint instance representations, we design the CLUE framework based on the contrastive learning-guided K-Means exemplar clustering algorithm. Contrastive K-Means is an unsupervised method that first learns a discriminative embedding space via a contrastive objective (attracting positives, repelling negatives under cosine distance) and then runs K-Means over those embeddings to produce centroid-based clusters. This method is particularly suitable for event induction, as the contrastive objective enforces intra-event cluster semantic coherence between texts of similar event types while maintaining inter-event cluster separation. Exemplar-based clustering further ensures cluster assignments by anchoring each cluster to a representative semantic prototype.

The corpus we use for clustering consists of original and augmented instances, where augmentations serve as positive pairs. The original instance serves as the anchor, and negative samples are sampled in-batch from instances different from the original instance. As shown in Eq~\eqref{eq:loss}, we optimize a contrastive triplet margin loss using cosine distance between each instance embedding and its corresponding exemplar. The exemplar in our case is the centroid or prototype instance that best represents the semantic concept of the event cluster. 

\subsubsection*{Choice of k}
\label{sec:K}
The value of \textit{k} is fundamental to the K-Means algorithm, and in real-world scenarios, it is difficult to predetermine the optimal value because the true number of events mentioned in the corpus is unknown. A larger \textit{k} would incur more LLM API calls to name the clusters and, similarly, more computational resources to merge semantically similar clusters. On the other hand, a smaller \textit{k} (less than the number of actual event clusters) would mean that many distinct event types would get merged under broad clusters. Prior work~\citep{shen-etal-2021-corpus} had arbitrarily chosen a fixed value of \textit{k} (=100). Certain strategies, such as the elbow method and silhouette score, are computationally expensive and may not translate well to the event-type induction task.  In CLUE, the \textit{k} value is approximated as 1.5 times the actual number of event types of the corpus. We have also explored different values of k and further discussion of \textit{k} is provided in Appendix~\ref{app:k_study}.   

\subsection{Event Type Naming}
Poorly named event categories are detrimental to the event induction task. Meaningful, semantically relevant event-type names help specify the semantic concept expressed by the cluster and are essential for downstream event extraction tasks~\citep{cao-etal-2023-event}. 

The framework not only provides meaningful event type names to the clusters but also leverages the initial event type names to refine and consolidate the event types. For this task, we use synthetic supervision with an LLM.

\noindent\textbf{Prompt template for event type naming: } LLMs have seen recent widespread usage in a vast array of NLP tasks and possess few-shot learning capabilities~\citep{llms_fewshot}. LLMs are extremely sensitive to the style and design template of the prompts used. In-context learning has been shown to help LLMs generate interpretable event type names for event clusters~\citep{cao-etal-2023-event, xu-etal-2024-ceo}. 

CLUE uses a prompt template specific to the event-type naming task and leverages the in-context learning principle. The prompt explains the textual content of the induced clusters to the LLM and uses the five sentences closest to the centroid of the event cluster as in-context examples. The objective of the event-naming task is to assign an event type name that serves as an abstraction of the semantic concept expressed by the event cluster, and the five sentences closest to the centroid best represent the abstract semantic concept of the event cluster. The five sentences are also accompanied by the predicate-object pairs extracted from them, along with the gold event type labels. Additionally, we provide examples of event type names (e.g., Conflict: Attack).

\noindent\textbf{Fine-grained definition of the event name:} In addition to the event type naming process, the LLM provides a fine-grained definition of the event type it has discovered. The fine-grained definition is used in the event type refinement step, as described in the next section. To infer event type names for our generated clusters, an LLM (GPT-4 Turbo) is used. The specific prompt templates used for this step are provided in Appendix~\ref {app:C} and on our GitHub.

\subsection{Event Type Refinement}
After the initial event cluster creation and naming, some induced event clusters may be semantically similar or too granular, and thus redundant as top-level event types. Semantically similar event types generally have similar event triggers, event type names, and event arguments. The same is true for granular event types, whose semantic concepts can already be represented by a top-level event type. In a manually supervised setting, similar or overly granular event types are merged or pruned by an annotator. However, in an unsupervised setting, it is difficult to determine which event types should be merged or consolidated. In an automated setting, the naive approach would be to iterate over each instance in the named event clusters and merge clusters with high textual overlap, but such a process is computationally expensive.

Instead, it is computationally cheaper to simply merge redundant event clusters using the semantic similarity of their event type names and fine-grained definitions. A similar method has been utilized for open-domain semantic event evaluation tasks~\citep{lu-etal-2025-seoe}. Semantic evaluation is deemed superior to token-level matching, as similar event type names (e.g., Conflict: Attack and Battle: War) may differ in tokens but can refer to the same event. The textual similarity model used for the threshold is a BGE-embedding-based model~\citep{chen-etal-2024-m3}. All types of events whose textual similarity is above a certain threshold (\textit{t} = 0.85) are merged and consolidated. The merged event clusters are randomly assigned an event name from the original event type names, and their instances are combined into one single group. In addition to refined event types, the event type labels also serve as top-level event type annotations for each sentence in our corpus. 

\section{Experiment}
\begin{table}[t]
\centering
\setlength{\tabcolsep}{4pt}
\renewcommand{\arraystretch}{1.15}
\resizebox{\columnwidth}{!}{%
\begin{tabular}{@{} l l r r r r @{}}
\toprule
\textbf{Dataset} & \textbf{Domain} & \textbf{\#Docs} & \textbf{\#Mentions} 
& \textbf{\makecell{\#Event\\Types}} 
& \textbf{\makecell{\#Top\\Event Types}} \\
\midrule
ACE   & News    & 599     & 5{,}349   & 33  & 8  \\
MAVEN & General & 4{,}480 & 118{,}732 & 168 & 5  \\
RAMS  & News    & 3{,}993 & 9{,}124   & 139 & 12 \\
\bottomrule
\end{tabular}%
}
\caption{Statistics of studied event datasets.}
\label{tab:tab1}
\end{table}

\begin{table*}[t]
\centering
\setlength{\tabcolsep}{7pt}
\resizebox{\textwidth}{!}{%
\begin{tabular}{lcccc|cccc|cccc}
\hline
\textbf{Method} &
\multicolumn{4}{c|}{\textbf{ACE}} &
\multicolumn{4}{c|}{\textbf{MAVEN}} &
\multicolumn{4}{c}{\textbf{RAMS}} \\
\cline{2-5}\cline{6-9}\cline{10-13}
& \textbf{P} $\uparrow$  & \textbf{R} $\uparrow$ & \textbf{F1} $\uparrow$  & \textbf{ARI} $\uparrow$
& \textbf{P} $\uparrow$  & \textbf{R} $\uparrow$  & \textbf{F1} $\uparrow$  & \textbf{ARI} $\uparrow$
& \textbf{P}  $\uparrow$ & \textbf{R} $\uparrow$  & \textbf{F1} $\uparrow$  & \textbf{ARI} $\uparrow$  \\
\hline
ETypeClus & 0.180 & 0.013 & 0.020 & 0.26 & 0.043 & 0.043 & 0.040 & 0.11 & 0.010 & 0.010 & 0.010 & 0.19 \\
CEO   & 0.180 & 0.360 & 0.240 & 0.28 & 0.131 & 0.172 & 0.150 & 0.28 & 0.150 & 0.280 & 0.200 & 0.25 \\
\hline
CLUE-GPT-J & 0.350 & 0.290 & 0.320 $\pm$ .000 & 0.29 $\pm$ .000 & 0.170 & 0.200 & 0.180 $\pm$ .000 & 0.33 $\pm$ .000 & 0.251 & 0.252 & 0.251 $\pm$ .000 & 0.62 $\pm$ .000 \\
CLUE-ICL-gold & 0.459 & 0.383 & 0.417 $\pm$ .014 & 0.31 $\pm$ .01 & 0.398 & 0.265 & 0.320 $\pm$ .018 & 0.50 $\pm$ .01 & 0.260 & 0.260 & 0.260 $\pm$ .010 & 0.62 $\pm$ .01 \\
\hline
\end{tabular}%
}
\caption{Performance comparison across datasets on sentence-level micro-F1 statistics and ARI metric (higher is better). CLUE-GPT-J utilizes GPT-J model for naming. CLUE-ICL-gold uses GPT-4 Turbo for naming. Mean and s.d.\ values are presented.}
\label{tab:tab2}
\end{table*}

\begin{table*}[t]
\centering
\setlength{\tabcolsep}{4pt}
\resizebox{\textwidth}{!}{%
\begin{tabular}{@{}lccccccccc@{}}
    \toprule
    & \multicolumn{3}{c}{\textbf{ACE}}
    & \multicolumn{3}{c}{\textbf{MAVEN}}
    & \multicolumn{3}{c}{\textbf{RAMS}} \\
    \cmidrule(lr){2-4} \cmidrule(lr){5-7} \cmidrule(lr){8-10}
    \textbf{Method}
    & \textbf{Sim dist} $\uparrow$ & \textbf{ROUGE-L} $\uparrow$ & \textbf{BERTScore} $\uparrow$
    & \textbf{Sim dist} $\uparrow$ & \textbf{ROUGE-L} $\uparrow$ & \textbf{BERTScore} $\uparrow$
    & \textbf{Sim dist} $\uparrow$ & \textbf{ROUGE-L} $\uparrow$ & \textbf{BERTScore} $\uparrow$ \\
    \midrule
    most frequent & .508 & .167 & .869 & .466 & .043 & .836 & .448 & .041 & .849 \\
    tf-idf        & .505 & .184 & .869 & .464 & .041 & .835 & .447 & .038 & .849 \\
    topicrank     & .437 & .024 & .824 & .380 & .000 & .721 & .413 & .006 & .817 \\
    textrank      & .418 & .035 & .813 & .376 & .000 & .724 & .399 & .016 & .811 \\
    keybert       & .462 & .072 & .838 & .427 & .000 & .795 & .425 & .014 & .830 \\
    WordNet       & .438 & .055 & .827 & .418 & .006 & .814 & .411 & .003 & .825 \\
    RoBERTa-large & .510 & .191 & .871 & .462 & .041 & .838 & .440 & .027 & .842 \\
    CEO           & .513 & .210 & .880 & .466 & .051 & .840 & .466 & .086 & .851 \\
    \midrule
    \textbf{CLUE-ICL-gold} & .949 $\pm$ .001 & .882 $\pm$ .001 & .979 $\pm$ .001
                  & .743 $\pm$ .001 & .342 $\pm$ .001 & .921 $\pm$ .001
                  & .808 $\pm$ .001 & .353 $\pm$ .001 & .928 $\pm$ .001 \\
    \textbf{CLUE-ZS} & .744 $\pm$ .001 & .358 $\pm$ .001 & .941 $\pm$ .001
                     & .727 $\pm$ .001 & .276 $\pm$ .001 & .898 $\pm$ .001
                     & .710 $\pm$ .001 & .203 $\pm$ .001 & .881 $\pm$ .001 \\
    \bottomrule
\end{tabular}%
}
\caption{
    Event Type Naming Quality on ACE, MAVEN, and RAMS using Sim dist, ROUGE-L, and BERTScore (higher is better). CEO baselines are from the original paper. CLUE-ICL-gold uses in-context learning for naming; CLUE-ZS names clusters zero-shot without access to any gold labels. Mean and s.d. values are presented.
}
\label{tab:tab3}
\end{table*}

\subsection{Datasets}
The standard datasets used by previous baseline methods have been selected to evaluate our framework. 1) \textbf{ACE} (Automatic Content Extraction)~\citep{doddington-etal-2004-automatic} is the most widely used news-wire domain closed-source dataset and was released by the Linguistic Data Consortium in 2005. 2) \textbf{MAVEN} (Massive Event detection)~\citep{wang-etal-2020-maven}  dataset is an open-source collection of Wikipedia articles. 3) \textbf{RAMS} (Roles Across Multiple Sentences)~\citep{ebner-etal-2020-multi} is an open-source news source dataset built using the AIDA ontology. All datasets used in this work are in English. The training set of the dataset is utilized for the evaluation. Detailed statistics of the dataset are provided in Table~\ref{tab:tab1}. 

\subsection{Baselines}
The performance of our framework is evaluated on the event type induction task against prior work: 1) \textbf{ETypeClus}~\citep{shen-etal-2021-corpus} and 2) \textbf{CEO}~\citep{xu-etal-2024-ceo}. We compare the quality of the named event clusters provided by CLUE against the clusters induced by EtypeClus and CEO. ETypeClus was the seminal work that introduced the unsupervised event type induction task. Since EtypeClus induced unnamed clusters, we named the clusters produced by the method using our naming methodology. The CEO method employs hierarchical event type ontology induction and names the induced clusters. For the event type naming task, the chosen evaluation baseline is the CEO method.

We note that while several methods, such as HALTON~\citep{cao-etal-2023-event}, include event-type naming evaluations, their overall task is event ontology completion. As our event type induction task uses an unlabeled raw corpus that does not include any seen event types, several weakly supervised or semi-supervised methods are omitted. 

\subsection{Evaluation Metrics}
The evaluation metrics are chosen to assess the quality of the induced event clusters and the event names provided by our framework. 

\subsubsection{\textnormal{\bfseries{Evaluating Event Cluster Quality}}}
We evaluate based on the top-level event type label at the sentence level. Our evaluation pipeline consists of the following: 1) Mapping the predicted major event type labels of the sentences in the merged clusters and aligning them with the ground truth labels using the BGE-embedding similarity model. A semantic evaluation approach is adopted where, if a predicted event type label is semantically similar to a ground truth type, we map that prediction label to the ground truth event type name~\citep{lu-etal-2025-seoe}. 2) For sentences that do not have any ground truth event mentions, we filter out those sentences. 3) For each sentence in our corpus, the prediction set of aligned event type labels is evaluated using the ground truth label set.  The micro-F1 and ARI~\citep{ari} are computed using the predicted and ground-truth event type label sets. 

\subsubsection{\textnormal{\bfseries{Evaluating Event Type Naming}}}
Our event type name evaluation compares the predicted event type names of the final event clusters with the ground truth event type labels in the dataset. As presented in the \textbf{CEO} baseline, the following metrics are used to infer the event name quality of the event clusters: 1) \textit{ Sim dist}~\citep{xu-etal-2024-ceo}; 2) \textit{ Rouge-L}~\citep{lin-2004-rouge}; and 3) \textit{ BERTScore}~\citep{bert-score}.

\subsection{Setup}
\label{sec:setup}
CLUE uses the \textbf{TextEE}~\citep{huang-etal-2024-textee} library to pre-process the datasets. The training set of the datasets is used for the experiments. P-O pairs are extracted using the Stanford CoreNLP toolkit~\citep{angeli-etal-2015-leveraging}. Stop-words and common verbs are filtered out. Each pair is combined with its sentence to form a joint instance, assigned a unique identifier, and encoded using BERT. Instances are augmented via masked language modeling (MLM), with up to five augmentations per instance. During augmentation, we filter out cases that result in errors and limit the context window length to ensure our representation does not exceed BERT's maximum token limit. The original and augmented instances are then clustered using a contrastive loss-based K-Means exemplar algorithm, with $k$ set to approximately 1.5 times the number of event types (i.e., $k=50$ for ACE, $252$ for MAVEN, and $209$ for RAMS). Training is performed for 5 epochs with a fixed random seed. Event clusters are named using GPT-4 Turbo and refined using the BGE embedding-based semantic similarity metric with a threshold of $t=0.85$. We run CLUE 5 times with different random seeds and report the mean and standard deviation. Additional implementation details are provided in the Appendix~\ref{app:A}.

\subsection{Results and Discussion}
\label{sec:results}
The effectiveness of our framework is demonstrated in Table~\ref{tab:tab2} and Table~\ref{tab:tab3}. On ACE, CLUE achieves a micro-F1 score of 41.7\%, compared to CEO's 24.0\%, resulting in a 17.7\% improvement, while the ARI increases by 3\% from 28\% to 31\%. On MAVEN, CLUE obtains a micro-F1 of 32\%, a 17\% relative improvement over CEO's 15\%, and the ARI improves from 28\% to 50\%, reflecting a 22\% improvement. For RAMS, CLUE achieves a micro-F1 of 26\%, a 6\% relative improvement over CEO's 20\%, while the ARI increases from 25\% to 62\%, reflecting a 37\% improvement.

CLUE identifies all 8 top-level event types in ACE, all 5 top-level event types in MAVEN, and all 12 top-level event types in RAMS. The event names obtained via prompting also outperform the existing state-of-the-art baseline methods as shown in Table~\ref{tab:tab3}. CLUE outperforms existing baselines in both the in-context learning (CLUE-ICL-gold) and zero-shot (CLUE-ZS) settings. Further evaluations and ablations are provided in Appendix~\ref{app:D}.

\subsection{Ablations}

\subsubsection{Effectiveness of the representation and augmentation}
\label{sec:represent}
An ablation study is conducted to evaluate the effectiveness of the representation and the utility of the augmentations using three settings. In the first setting, only the P-O pair is considered for the K-Means clustering algorithm while keeping the augmentations of the predicate verb. In the second setting, we use the sentence as input to the K-Means clustering algorithm, but we keep the augmented instances of the predicate verb. In the third setting, we use our joint instance representation of the predicate verb-object pair with the sentence, but we do not perform any augmentations. 

Experimental results presented in Table~\ref{tab:tab4} show that removing any of the components in CLUE leads to a decrease in performance. The use of only the P-O pair is less effective than the joint representation, which incorporates both the P-O pair and the sentence. 
\begin{table}[t]
\small
\resizebox{\columnwidth}{!}{%
\begin{tabular}{@{}lccc@{}}
\toprule
\textbf{Method} & \textbf{Prec.} $\uparrow$ & \textbf{Rec.} $\uparrow$ & \textbf{F1} $\uparrow$ \\
\midrule
\textbf{CLUE} & \textbf{0.47} & \textbf{0.39} & \textbf{0.43} \\
Cluster with P-O pairs only   & 0.34 & 0.36 & 0.35 \\
Cluster with sentences only   & 0.40 & 0.35 & 0.37 \\
Cluster without augmentations & 0.32 & 0.27 & 0.29 \\
\bottomrule
\end{tabular}%
}
\caption{Ablation on the Representation.}
\label{tab:tab4}
\end{table}
\subsubsection{Effectiveness of Margin Loss}
\label{sec:loss}
As shown in Table~\ref{tab:tab5}, the ablation study on the loss function shows that margin loss is better than InfoNCE loss regarding precision and overall F1-score. The clustering aims to induce semantically coherent clusters that are neither overly granular nor too broad. Ideally, the clustering algorithm should assign instances to clusters to maximize precision and capture all semantically similar instances while discarding dissimilar ones. Since the number of dissimilar instances far outnumbers those that are similar in a dataset, we prefer the loss function that maximizes precision and the overall F1-score. Hence, we choose the margin loss function for the K-Means clustering algorithm.
\begin{table}[h]
\centering
\small
\setlength{\tabcolsep}{4pt}
\begin{tabular}{lccc}
\toprule
\textbf{Loss Function} & \textbf{Prec.}  $\uparrow$ & \textbf{Rec.} $\uparrow$  & \textbf{F1} $\uparrow$  \\
\midrule
InfoNCE & 0.44 & \textbf{0.41} & 0.42 \\
Margin  & \textbf{0.47} & {0.40} & \textbf{0.43} \\
\bottomrule
\end{tabular}
\caption{F1-scores for InfoNCE vs.\ Margin loss}
\label{tab:tab5}
\end{table}

\subsection{Case Study on Cluster Quality and Naming}
\label{sec:case}
Clustering is an under-specified problem, and real-world applications may lack ground-truth labels. To judge the quality and nature of the event clusters, a trained undergraduate annotator is assigned to assess the cluster coherence before and after merging, as well as the alignment of cluster names.

The annotator rates cluster coherence on a 1–5 scale, where 1 indicates low coherence, and 5 indicates high coherence. The annotator is provided with the cluster ID, five sentences closest to the centroid, the majority ground-truth label based on the five sentences, their corresponding P–O pairs, and the LLM-assigned event name. Based on this information, the annotator assesses the semantic similarity and coherence of the cluster, both before and after merging. The final score is computed as the average coherence across clusters.

The annotator is also tasked with judging the alignment of the event type names. Here, the LLM's complete event type name is manually evaluated. The LLM event type naming may result in names that are not part of the ACE event types, and some clusters can be associated with novel event types (e.g., Sports: Game). The annotator is asked to rate event type names as Correct if the LLM correctly reflects the event type of that cluster, Wrong if the LLM provides an event type name that differs from the ground truth event types, or Novel if the LLM provides a name not present in the list of ACE event type names.  

The results of the case study are presented in Table~\ref{tab:tab6}. Manual evaluation shows that merging improves cluster coherence, with a 0.27 increase in the mean score while maintaining comparable event naming quality. The LLM can also identify novel event types, thereby aiding humans in the annotation process.
\begin{table}[t]
\centering
\small
\setlength{\tabcolsep}{3pt}
\renewcommand{\arraystretch}{1.05}
\resizebox{\columnwidth}{!}{%
\begin{tabular}{l c c c c c}
\toprule
\textbf{Cluster} & \textbf{Total} & \textbf{Mean coherence} & \multicolumn{3}{c}{\textbf{Naming quality}} \\
\textbf{type} & \textbf{clusters} & \textbf{score (1--5)} & \textbf{Correct} & \textbf{Wrong} & \textbf{Novel} \\
\midrule
Pre-merge  & 50 & 4.06 & 33 & 11 & 6 \\
Post-merge & 18 & 4.33 & 12 & 4  & 2 \\
\bottomrule
\end{tabular}%
}
\caption{Case Study Results on ACE clusters before and after name-based consolidation. Event Naming Quality (total event type) is reported as counts of \textit{Correct/Wrong/Novel}.}
\label{tab:tab6}
\end{table}

\subsection{Analysis of the framework}
CLUE is a modular framework and can be implemented piecemeal. The k-value, semantic threshold, and event-naming prompt templates can be adjusted based on user criteria. For the stated purpose of generating new event datasets, CLUE can create annotated event datasets and is also well-suited for human-in-the-loop systems. Human annotators often overlook rare, long-tail, or domain-specific event types, particularly in large-scale corpora. In fact, the novel event types discovered by CLUE can help enrich existing datasets by expanding their coverage and removing manual bias. Beyond the event clustering task, CLUE can be extended to related unsupervised semantic clustering tasks, such as topic-like grouping of sentences, and in situations where polysemy disambiguation is required. 

\section{Conclusion}
The paper proposes the CLUE framework for inducing the top-level event ontology from raw text without external verb resources. It introduces a joint instance representation to handle verb polysemy, applies augmentation with contrastive loss-guided K-Means clustering to form coherent event clusters, and assigns interpretable cluster names. CLUE refines event types through semantic similarity and produces sentence-level event annotations, enabling thorough semantic evaluation. Experiments show that CLUE effectively induces event types from unlabeled corpora and outperforms prior baselines while keeping annotation costs low.

\section{Acknowledgements} The work is supported in part by NSF-CAREER \#2237831 and USDA-NIFA \#2024-08585.

\section*{Limitations}

The CLUE framework can produce variable output due to differences in the clusters produced by the K-Means clustering algorithm. Also, K-Means clustering assumes that clusters are spherical and uniform in size. However, K-Means may not be an optimal fit for data with skewed distributions. We have provided ablations on the role of k in our algorithm. 

CLUE only induces the flat or top-level ontology and does not produce hierarchical event types. Our method can be extended to hierarchical event type induction by recursively applying the method to the initial event clusters. However, this naive extension may lead to error propagation issues. We leave the task of hierarchical event type induction to future work.

Also, our preliminary focus is not on prompt engineering; thus, a generic prompt has been utilized in our work. We did not explore other strategies that may further improve the event type naming and event type refinement tasks.


\appendix

\section{Implementation Details}
\label{app:A}
This section presents the hyperparameters and prompt settings for the CLUE framework for event cluster induction and event cluster naming, respectively.

In Table~\ref{tab:tab11}, the hyperparameter values used in the experimental setup for the CLUE framework are presented. 

\begin{table}[h]
\centering
\small
\begin{tabular}{lc}
\hline
\textbf{Hyperparameter} & \textbf{Value} \\
\hline
Epochs        & 5 \\
Batch size    & 16 \\
Learning rate & $1 \times 10^{-5}$ \\
Weight decay  & $1 \times 10^{-4}$ \\
Margin        & 0.75 \\
\hline
\end{tabular}
\caption{CLUE hyperparameter values used in all experiments.}
\label{tab:tab11}
\end{table}

In Table~\ref{tab:tab7}, the prompt parameters used for event naming and for obtaining the fine-grained definitions of the event names are reported.

\begin{table}[h]
\centering
\resizebox{\columnwidth}{!}{
\begin{tabular}{lc}
\toprule
\textbf{Parameter} & \textbf{Value} \\
\midrule
LLM Model & GPT-4 Turbo \\
Temperature & 0.3 \\
Number of Examples & 5 \\
Example Selection & Centroid-based closest 5 sentences \\
Output Format & XML tags (\texttt{<EventType>}) \\
\bottomrule
\end{tabular}}
\caption{Prompt settings for LLM-Based Event Naming}
\label{tab:tab7}
\end{table}

\section{Running Environment}
\label{app:B}

The experimental setup and evaluations were carried out on a machine with a 16-core CPU, an RTX 4090 GPU, and 96 GB of RAM, as well as on a multi-GPU system with A100 GPUs. Furthermore, an API was used to obtain the LLM outputs. The total GPU hours for running the CLUE framework on the three datasets were approximately three hours. The cost for LLM API calls (via Open Router and OpenAI/Claude API) was approximately  50 dollars.

The following Python libraries have been used which include Stanford CoreNLP for linguistic preprocessing,
BERT (\texttt{bert-base-uncased}) from Hugging Face Transformers with
PyTorch for encoding the joint instance, and also \texttt{scikit-learn} (with optional
FAISS) for $K$-Means clustering. In terms of LLMs, we used GPT-4 Turbo via the OpenRouter API
for event naming and definition generation. For evaluations and semantic matching, we utilize
\texttt{BAAI/bge-base-en-v1.5} via FlagEmbedding for semantic
similarity, and the \texttt{rouge\_score} and \texttt{bert\_score}
packages for ROUGE-L and BERTScore evaluation.

\section{Prompts used for Naming and Description}
\label{app:C}

The prompt templates used for naming the event clusters and for obtaining the event type name description are provided in this section. 

\begin{figure}[h]
\centering
\includegraphics[width=0.75\columnwidth]{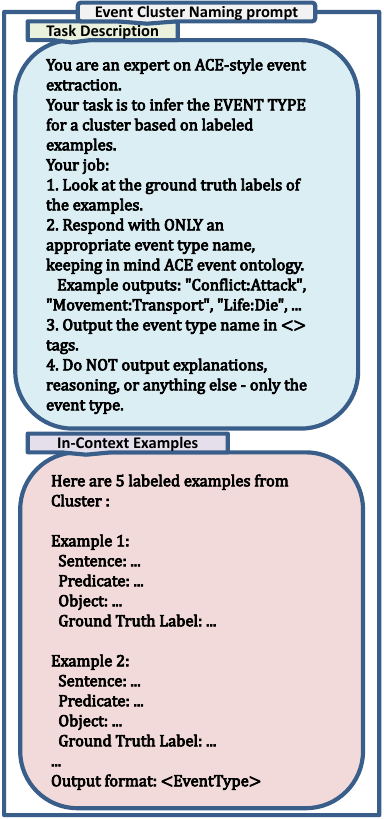}  
\caption{Prompt template for LLM-based event naming.}
\label{fig:fig3}
\end{figure}

We provide clear instructions to the LLM for the event naming task. We start by clearly defining and explaining the event cluster naming to the LLM. We then provide in-context examples consisting of five sentences closest to the centroid of the event cluster to be named. The examples include the original sentence, with the predicate verb lemma on the next line, followed by the object lemma. Then we provide the ground-truth labels for the example sentences. Finally, we ask the LLM to provide the output in XML <> tags so we can easily extract the event type name. We note that the event names are constrained by the specific dataset used, and the in-context examples might not be present in a truly unsupervised setting. The prompt for event naming is provided in Figure~\ref{fig:fig3}.

\begin{figure}[h]
\centering
\includegraphics[width=0.75\columnwidth]{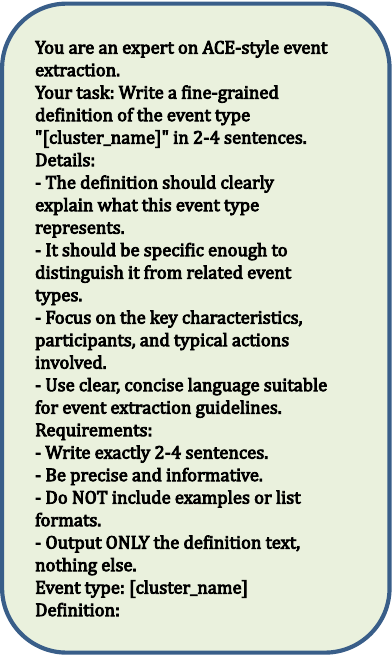}  
\caption{Prompt template for LLM-based event type definition.}
\label{fig:fig4}
\end{figure}

The prompt for the event description is provided in Figure~\ref{fig:fig4}.

\section{Additional Evaluation and Ablations}
\label{app:D}

\subsection{Evaluation on additional event datasets}
\label{app:dataset_add}

We evaluate our framework on additional datasets and present the event annotation and naming quality for these datasets.

The additional datasets on which we evaluated our framework are: 1) CASIE~\cite{casie}, 2) GENIA~\citep{kim-etal-2013-genia}, 3) M2E2~\citep{li-etal-2020-cross}, 4) PHEE~\citep{sun-etal-2022-phee}, 5) WikiEvents~\citep{li-etal-2021-document}.

The statistics of the additional datasets are in Table~\ref{tab:tab8}.

\begin{table}[h]
\centering
\scriptsize
\setlength{\tabcolsep}{7pt}
\renewcommand{\arraystretch}{1.1}
\resizebox{\columnwidth}{!}{
\begin{tabular}{llrrr}
\toprule
\textbf{Dataset} & \textbf{Domain} & \textbf{\#Docs} & \textbf{\#Mentions} & \textbf{\#Event Types} \\
\midrule
CASIE            & Cybersecurity     & 999   & 8,469  & 5  \\
GENIA       & Biomedical        & 20    & 6,001  & 13 \\
M2E2             & Multimedia        & 6,013 & 1,105  & 8  \\
PHEE             & Pharmacovigilance & 4,827 & 5,019  & 2  \\
WikiEvents       & Wikipedia         & 245   & 3,932  & 50 \\
\bottomrule
\end{tabular}}
\caption{Statistics of additional datasets used in our experiments.}
\label{tab:tab8}
\end{table}

In Table~\ref{tab:tab9} we report the micro-F1 metric and the ARI values for the datasets. We note that the low recall values for GENIA and PHEE are due to the fact that our method names the top-level event types for a dataset. Both datasets have hierarchical event types that might not align with the event names generated by our framework.

\begin{table}[h]
\centering
\small
\setlength{\tabcolsep}{4pt}
\begin{tabular}{lcccc}
\toprule
\textbf{Dataset} & \textbf{Prec.} & \textbf{Rec.} & \textbf{F1} & \textbf{ARI} \\
\midrule
CASIE & 0.628 & 0.549 & 0.586 & 0.550 \\
GENIA & 0.610 & 0.180 & 0.278 & 0.579 \\
M2E2 & 0.548 & 0.495 & 0.520 & 0.298 \\
PHEE & 0.863 & 0.845 & 0.854 & 0.297 \\
WIKIEVENTS & 0.730 & 0.235 & 0.355 & 0.490 \\
\bottomrule
\end{tabular}
\caption{Additional Dataset Evaluations on micro-F1 and ARI}
\label{tab:tab9}
\end{table}

In Table~\ref{tab:tab10}, we report the event naming metrics, specifically the Sim dist, ROUGE-L, and BERTScore metrics for the datasets.

\begin{table}[h]
\centering
\resizebox{\columnwidth}{!}{
\begin{tabular}{lccc}
\toprule
\textbf{Dataset} & \textbf{Sim\_dist} & \textbf{ROUGE-L} & \textbf{BERT Score} \\
\midrule
CASIE & 0.840 & 0.430 & 0.880 \\
GENIA & 0.980 & 0.970 & 0.990 \\
M2E2 & 0.867 & 0.738 & 0.929 \\
PHEE & 1.000 & 1.000 & 1.000 \\
WIKIEVENTS & 0.900 & 0.740 & 0.950 \\
\bottomrule
\end{tabular}}
\caption{Additional Evaluation on Event Naming}
\label{tab:tab10}
\end{table}

\begin{table}[!ht]
\centering
\small
\setlength{\tabcolsep}{8pt}
\renewcommand{\arraystretch}{1.1}
\begin{tabular}{lrr}
\toprule
\textbf{Dataset} & \textbf{Total Event Types} & \textbf{$k$} \\
\midrule
ACE      & 33  & 50  \\
MAVEN        & 168 & 252 \\
RAMS         & 139 & 209 \\
CASIE        & 5   & 8   \\
M2E2         & 8   & 12  \\
PHEE         & 2   & 3   \\
WIKIEVENTS   & 50  & 76 \\
GENIA        & 13  & 20  \\
\bottomrule
\end{tabular}
\caption{Number of Total Event Types and value of $k$ (number of clusters) used for each dataset.}
\label{tab:tab12}
\end{table}

\subsection{Performance of various LLMs on the Event Naming Task}
\label{app:llm}

\begin{table*}[!ht]
\centering
\small
\begin{tabular}{lccc}
\hline
\textbf{Model} & \textbf{Sim\_dist} $\uparrow$ & \textbf{ROUGE-L} $\uparrow$ & \textbf{BERT Score} $\uparrow$ \\
\hline
GPT-5.4 & \textbf{0.921} & \textbf{0.791} & 0.975 \\
Gemini 3.1 Pro & 0.904 & 0.750 & \textbf{0.981} \\
GPT-4 Turbo & 0.900 & 0.731 & 0.970 \\
Deepseek-v3.2 & 0.871 & 0.661 & 0.971 \\
Qwen-3 VL 32B-Instruct & 0.834 & 0.594 & 0.956 \\
Claude Opus-4.6 & 0.832 & 0.592 & 0.954 \\
Ministral 3 8B & 0.783 & 0.497 & 0.891 \\
Llama-3.1 70B-Instruct & 0.773 & 0.406 & 0.932 \\
Llama-3.1 8B-Instruct & 0.771 & 0.408 & 0.934 \\
Ministral 3 32B & 0.769 & 0.386 & 0.929 \\
Qwen-3 VL 8B-Instruct & 0.766 & 0.390 & 0.945 \\
Phi-4 & 0.743 & 0.392 & 0.937 \\
Llama-3.2 3B-Instruct & 0.731 & 0.310 & 0.923 \\
\hline
\end{tabular}
\caption{Performance comparison across LLMs on the event naming task. Sim\_dist, ROUGE-L, and BERT Score are reported (higher is better). Best results per column are in \textbf{bold}.}
\label{tab:model_comparison}
\end{table*}

\begin{table}[!ht]
\centering
\small
\begin{tabular}{lcccc}
\hline
\textbf{$k$} & \textbf{Factor} & \textbf{P} & \textbf{R} & \textbf{F1} \\
\hline
33 & 1.00 & 0.450 & 0.370 & 0.406 \\
42 & 1.25 & 0.501 & 0.401 & 0.445 \\
50 & 1.50 & 0.505 & 0.415 & 0.456 \\
66 & 2.00 & 0.483 & 0.400 & 0.438 \\
83 & 2.50 & 0.497 & 0.410 & 0.450 \\
\hline
\end{tabular}
\caption{Ablation results across different values of $k$ on ACE dataset, which has 33 event types. The Factor column denotes $k*33$. Precision (P), Recall (R), and F1 scores are reported. Best results are in \textbf{bold}.}
\label{tab:k_ablation}
\end{table}

We showcase the performance of various LLMs on the event type naming task in Table~\ref{tab:model_comparison}. We observe that closed-source models perform better than open-source models. Additionally, the latest frontier models are more accurate than the older models. However, it should also be noted that the closed-source frontier models are more expensive in terms of API costs. Our framework can use a variety of models, including open-source, lower-parameter LLMs, and still demonstrate state-of-the-art performance on the event naming task.

\subsection{Annotation Quality and K}
\label{app:k_study}

We report the results of an ablation study conducted on the ACE dataset that uses varying k values for the K-Means clustering algorithm. ACE has 33 total event types and 8 major event types. We perform the ablation by increasing the \textit{k} value from 33 (where we used a multiplicative factor of 1 compared to the actual 33 event types) to 83 (a multiplicative factor of 2.5).  We report the results in Table~\ref{tab:k_ablation}. As observed in the table, a \textit{k} value greater than 1.5 provides little benefit in terms of improving event annotation quality. Additionally, we incur more expenses in LLM API costs as a greater number of clusters have to be named and defined. Thus, an optimal value of \textit{k} is chosen as 1.5 times the actual number of clusters in the dataset. In an open-domain scenario, the user can opt to choose a k that is 1.5 times their desired number of event types. A very large \textit{k} would be detrimental to keeping costs low in terms of API calls and computational resource expenditure.

\subsection{Semantic threshold (t) for merging and coverage of event types}
\label{app:t_ablatiom}

We also perform an ablation study on the semantic threshold value (t) for merging event type names. The study is conducted on the ACE dataset. The results are presented in table~\ref{tab:t_ablation}. It is observed that lowering the $t$ value merges clusters more aggressively and reduces the coverage of these major event types. Higher values of $t$ recover full coverage. However, choosing token-level matching of event type names over semantic matching results in a decrease in F1-score. Semantic matching allows merging semantically similar names that might express the same event but differ in exact wording. A strict token-level matching would treat semantically similar event types as distinct. As discussed previously, strict token-level matching results in event types like Conflict:Attack and Battle:War being treated as semantically distinct events. This problem is avoided by choosing the BGE-embedding threshold $t=0.85$.

\begin{table}[!ht]
\centering
\small
\begin{tabular}{lcc}
\hline
\textbf{Setting} & \makecell{\textbf{Coverage of}\\\textbf{Major Event Types}} & \textbf{Micro-F1} \\
\hline
$t=0.55$            & 1/8 & 0.04 \\
$t=0.70$            & 5/8 & 0.17 \\
$t=0.85$ (baseline) & 8/8 & \textbf{0.42} \\
Exact Match         & 8/8 & 0.37 \\
\hline
\end{tabular}
\caption{Ablation over the merge threshold $t$ on the ACE dataset, compared against exact-match merging. ACE has 8 major event types. Lowering $t$ merges clusters more aggressively and reduces the coverage of these major event types, while higher values of $t$ recover full coverage. Micro-F1 is aggregated over all evaluated sentences. Best result in \textbf{bold}.}
\label{tab:t_ablation}
\end{table}

\section{Annotation Cost Reduction based on sentence and token usage}
\label{app:token_usage}

\begin{table*}[!ht]
\centering
\small
\setlength{\tabcolsep}{6pt}
\renewcommand{\arraystretch}{1.2}
\resizebox{\textwidth}{!}{%
\begin{tabular}{lrrrrrr}
\toprule
\textbf{Dataset} & \makecell{\textbf{Unique Sentences}\\\textbf{(Filtered)}} & \makecell{\textbf{Total Token Count of}\\\textbf{Unique Sentences}} & \textbf{$k$} & \makecell{\textbf{Sentences sampled per}\\\textbf{cluster for Prompting}} & \makecell{\textbf{Total Sentences used}\\\textbf{for Prompting}} & \makecell{\textbf{Prompt Tokens used}\\\textbf{(incl. prompt text)}} \\
\midrule
ACE   & 6,676  & 144,259 & 50  & 5 & 250   & 21,355  \\
MAVEN & 25,235 & 614,713 & 252 & 5 & 1,280 & 188,521 \\
RAMS  & 7,417  & 991,029 & 209 & 5 & 1,045 & 282,787 \\
\bottomrule
\end{tabular}%
}
\caption{Token and sentence usage statistics per dataset. \textbf{Unique Sentences (Filtered)} is the number of unique sentences remaining after filtering, and \textbf{Total Token Count of Unique Sentences} is the total number of tokens across them. \textbf{$k$} is the number of clusters; CLUE samples 5 sentences from each cluster. \textbf{Prompt Tokens used} is the total number of tokens across all naming and definition prompts, including the full prompt text. Tokens are counted as whitespace-separated words.}
\label{tab:token_usage}
\end{table*}

A key goal of the CLUE framework is to reduce annotation costs for event type induction and event annotation. The joint P-O pair and sentence embedding design reduces manual annotation costs by eliminating the need for a domain-specific verb dictionary. The clustering and prompting strategy also aims to reduce the LLM API cost, which can be extremely expensive. As discussed earlier, we limit the number of clusters to 1.5 times the actual number of event types in the dataset. This reduces the cost of using the LLM for naming clusters without adversely affecting annotation quality. In Table~\ref {tab:token_usage}, we also show a token and sentence-level analysis of the  CLUE framework. The prompt tokens used for naming the clusters include the tokens from the 5 sampled sentences closest to the event cluster's centroid, as well as the instructions given to the LLM for the event cluster naming and definition task. As shown in the table, for the ACE dataset, CLUE uses only 250 of the 6,676 unique sentences. In terms of token usage, CLUE utilizes only 21,355 prompt tokens out of 144,259 total tokens. The same trend is observed at the sentence and token levels in the MAVEN and RAMS datasets. Thus, CLUE can induce event types and annotate the dataset using only a fraction of the total available tokens for LLM API calling. While CLUE uses the entire corpus of sentences for clustering, the cost of LLM API usage far exceeds that of running the clustering algorithm.

\section{Novel Event Names and Semantic Evaluation}
\label{app:F}

\begin{table*}[!ht]
\centering
\small
\setlength{\tabcolsep}{4pt}
\renewcommand{\arraystretch}{1.2}
\begin{tabular}{|p{2.2cm}|p{2.0cm}|p{2.4cm}|p{8.2cm}|}
\hline
\textbf{CLUE annotated Event Cluster Name} & \textbf{Ground Truth Event Type Name of majority of the sampled sentences} & \textbf{Human annotated Event Cluster Name} & \textbf{Sampled Sentences} \\
\hline
Government: Elect & Personnel: Elect & election, democrats/justice &
(1) On the other hand, I have yet to see any Democrats challenge incumbents for being fiscally irresponsible. \newline
(2) I think it's important to recognize that African Americans, though still overwhelmingly Democratic, comprise less of a monolith than they did 20--30 years ago. \newline
(3) Did you vote in the recent US Presidential election? \newline
(4) Our alliance to them is supporting people that want to have a free democratically elected gov't. \newline
(5) I actually can't recall it ever taking 2 weeks to ``count'' these many ballots, generally elections are called within 24 hours. \\
\hline
Conflict: Use-Conventional-Force & Conflict: Attack & nuclear, weapon/conflict &
(1) A senior lawmaker seen as close to the Kremlin ridiculed the United States Sunday for failing to find any weapons of mass destruction in Iraq. \newline
(2) A North Korean foreign ministry spokesman, quoted by the official Korean Central News Agency, made no reference to the reports that North Korea had admitted to possessing nuclear weapons. \newline
(3) Rodong called for a ``fair and equal'' footing in talks with the United States, blasting Washington for insisting that there would be ``no security of the system nor provision of rewards'' even if Pyongyang gave up its nuclear program. \newline
(4) The plant, based on Russian design and upgraded with U.S. technology, has been plagued by frequent, non-nuclear malfunctions since testing began in November 2000. \newline
(5) North Korea has never tested a nuclear weapon, though the CIA long ago estimated that it manufactured two crude devices in the late 1980s or early 1990s. \\
\hline
\end{tabular}
\caption{Example clusters showing CLUE-assigned label, ground truth event type name for the majority of the sampled sentences, human annotated cluster name, and the five sampled sentences.}
\label{tab:cluster_examples}
\end{table*}

As discussed earlier, the CLUE framework names and refines the event clusters. The event naming is done by prompting an LLM to provide the appropriate event name for the cluster. We apply constraints to the event cluster naming task by asking the LLM to name clusters based on the given dataset (e.g. ACE) and providing in-context examples. In an unsupervised setting, such constraints might be removed, and the event naming prompt template can be modified to ask the LLM to provide event type names. Despite our constraints, the LLM often provides novel event type names for the clusters that closely align with the actual events expressed in the cluster's sentences. While in some cases the actual major event label may not be a perfect match for the cluster, the CLUE-annotated event name is semantically closest to the event type expressed by the sampled sentences. This is illustrated in Table~\ref{tab:cluster_examples}. In the first row of the table, the sampled sentences are chosen from an induced cluster in the ACE corpus. The sentences represent the semantic concept of an election and can be expressed as the event of "getting a government elected" ( Government: Elect) or "getting a personnel/person elected" (Personnel: Elect). While both labels are semantically coherent for this cluster, a strict token-level matching would consider the CLUE annotation to be incorrect. Thus, only the semantic evaluation approach can appropriately judge the quality of the CLUE event naming. Also, the alternative CLUE event cluster name can be thought of as an additional label for the sentences. In the second row of the table, we see that the LLM has named the cluster Conflict: Use-Conventional-Force instead of Conflict: Attack. While our method focuses only on the major event type and does not produce hierarchical event clusters, we see that the CLUE method can also provide alternative minor sub-event type annotations for the corpus. In both cases, the CLUE annotation and the ground truth annotation are semantically similar.

\section{Additional Formulas and Definitions}
\label{app:E}

We define the unsupervised event type induction task as the process of discovering major event type labels without access to the ground truth labels for the clustering task. CLUE utilizes a contrastive K-Means exemplar clustering for the induction task without access to the ground truth event type labels. The loss function for CLUE is defined as:

\[
\mathcal{L}_{\text{CLUE}}
=
\mathcal{L}_{\text{triplet}}
+
\mathcal{L}_{\text{exemplar}}
\]
Here $\mathcal{L}_{\text{triplet}}$ is triplet loss, $\mathcal{L}_{\text{exemplar}}$ is exemplar loss.

The micro-F1 score is calculated as follows:
\begin{multline}
\text{Micro-F1} = \\
\frac{2 \cdot \sum_{i=1}^{N} \text{TP}_i}{2 \cdot \sum_{i=1}^{N} \text{TP}_i + \sum_{i=1}^{N} \text{FP}_i + \sum_{i=1}^{N} \text{FN}_i},
\end{multline}
where $N$ is the number of evaluated sentences. Since a sentence can have multiple top-level event types, for each sentence $i$
\begin{align*}
\text{TP}_i &= |\{y\}_i \cap \{\hat{y}\}_i|, \\
\text{FP}_i &= |\{\hat{y}\}_i \setminus \{y\}_i|, \\
\text{FN}_i &= |\{y\}_i \setminus \{\hat{y}\}_i|.
\end{align*}
where $\{y\}_i$ denotes the set of ground truth labels and $\{\hat{y}\}_i$ denotes the set of predicted labels.


We have employed a positive-label-only approach when computing the ARI metric~\citep {ari} for our framework. The ARI is considered for sentence-level event detection by evaluating each event type independently as a binary partition over all sentences in our corpus. We exclude cases in which both the gold standard and the predictions are absent for that event type.
\begin{equation}
\text{ARI} = \frac{\text{RI} - \mathbb{E}[\text{RI}]}{\max(\text{RI}) - \mathbb{E}[\text{RI}]},
\end{equation}
where the Rand Index (RI) is defined as $\text{RI} = \frac{\text{TP} + \text{TN}}{N_e}$, $N_e$ is the total number of instance pairs, and $\mathbb{E}[\text{RI}]$ is the expected value of the Rand Index under random partitioning.

The metrics for event name quality evaluation are: 1) \textit{ Sim dist}~\citep{xu-etal-2024-ceo} is defined to consider both the semantic and granular differences between the reference event name and the predicted event name; 2) \textit{ Rouge-L}~\citep{lin-2004-rouge}: measures the longest common subsequence (LCS) between a reference event type name and a predicted event type name; and 3) \textit{ BERTScore}~\citep{bert-score}: measures the similarity between a ground truth event name and a predicted event name using contextual embeddings.

\textit{Sim\_dist} is defined to consider both semantic similarity and granularity differences between each pair of reference $e_i^r$ and generated names $e_j^p$:
\begin{align}
\label{eq:simdist}
\mathrm{sim\_dist} &= \frac{1}{n_r \cdot n_p}\sum_{i,j}
\underbrace{\left(1 - \left|\frac{i}{n_r} - \frac{j}{n_p}\right|\right)}_{\text{granularity diff.}} \notag \\
&\quad \cdot \underbrace{\frac{\cos(\mathrm{emb}(e_i^r), \mathrm{emb}(e_j^p))+1}{2}}_{\text{semantic sim.}},
\end{align}
\noindent where $\mathrm{emb}$ is the phrase representation from SBERT~\citep{reimers-gurevych-2019-sentence}.

\textit{ROUGE-L}  is defined as follows:

\begin{align}
R_{\text{LCS}} &= \frac{\text{LCS}(X, Y)}{m}, \\
P_{\text{LCS}} &= \frac{\text{LCS}(X, Y)}{n}, \\
F_{\text{LCS}} &= \frac{(1 + \beta^2) \cdot R_{\text{LCS}} \cdot P_{\text{LCS}}}{R_{\text{LCS}} + \beta^2 \cdot P_{\text{LCS}}},
\end{align}

\noindent where $X$ is the reference event type name of length $m$, $Y$ is the predicted event type name of length $n$, $\text{LCS}(X, Y)$ is the length of their longest common subsequence, and $\beta$ controls the relative importance of recall versus precision.

\textit{BERTscore} is defined as follows:
\begin{align}
R_{\text{BERT}} &= \frac{1}{|x|} \sum_{x_i \in x} \max_{\hat{x}_j \in \hat{x}} \mathbf{x}_i^\top \hat{\mathbf{x}}_j, \\
P_{\text{BERT}} &= \frac{1}{|\hat{x}|} \sum_{\hat{x}_j \in \hat{x}} \max_{x_i \in x} \mathbf{x}_i^\top \hat{\mathbf{x}}_j, \\
F_{\text{BERT}} &= 2 \cdot \frac{P_{\text{BERT}} \cdot R_{\text{BERT}}}{P_{\text{BERT}} + R_{\text{BERT}}},
\end{align}

\noindent where $\mathbf{x}_i^\top \hat{\mathbf{x}}_j$ is the cosine similarity (inner product of normalized embeddings) between token embeddings $\mathbf{x}_i$ and $\hat{\mathbf{x}}_j$.

\section{AI Usage Disclosure}
\label{app:AI}
AI was used only for fixing grammar and coding bugs.


\begin{thebibliography}{49}
\providecommand{\natexlab}[1]{#1}

\bibitem[{Angeli et~al.(2015)Angeli, Johnson~Premkumar, and Manning}]{angeli-etal-2015-leveraging}
Gabor Angeli, Melvin~Jose Johnson~Premkumar, and Christopher~D. Manning. 2015.
\newblock \href {https://doi.org/10.3115/v1/P15-1034} {Leveraging linguistic structure for open domain information extraction}.
\newblock In \emph{Proceedings of the 53rd Annual Meeting of the Association for Computational Linguistics and the 7th International Joint Conference on Natural Language Processing (Volume 1: Long Papers)}, pages 344--354, Beijing, China. Association for Computational Linguistics.

\bibitem[{Bach(1986)}]{BachAlgebra}
Emmon Bach. 1986.
\newblock \href {https://doi.org/10.1007/BF00627432} {The algebra of events}.
\newblock \emph{Linguistics and Philosophy}, 9:5--16.

\bibitem[{Bonial et~al.(2016)Bonial, Tahmoush, Windisch~Brown, and Palmer}]{bonial-etal-2016-multimodal}
Claire Bonial, David Tahmoush, Susan Windisch~Brown, and Martha Palmer. 2016.
\newblock \href {https://doi.org/10.18653/v1/W16-1003} {Multimodal use of an upper-level event ontology}.
\newblock In \emph{Proceedings of the Fourth Workshop on Events}, pages 18--26, San Diego, California. Association for Computational Linguistics.

\bibitem[{Bronstein et~al.(2015)Bronstein, Dagan, Li, Ji, and Frank}]{bronstein-etal-2015-seed}
Ofer Bronstein, Ido Dagan, Qi~Li, Heng Ji, and Anette Frank. 2015.
\newblock \href {https://doi.org/10.3115/v1/P15-2061} {Seed-based event trigger labeling: How far can event descriptions get us?}
\newblock In \emph{Proceedings of the 53rd Annual Meeting of the Association for Computational Linguistics and the 7th International Joint Conference on Natural Language Processing (Volume 2: Short Papers)}, pages 372--376, Beijing, China. Association for Computational Linguistics.

\bibitem[{Brown et~al.(2017)Brown, Bonial, Obrst, and Palmer}]{brown-etal-2017-rich}
Susan Brown, Claire Bonial, Leo Obrst, and Martha Palmer. 2017.
\newblock \href {https://doi.org/10.18653/v1/W17-2712} {The rich event ontology}.
\newblock In \emph{Proceedings of the Events and Stories in the News Workshop}, pages 87--97, Vancouver, Canada. Association for Computational Linguistics.

\bibitem[{Brown et~al.(2020)Brown, Mann, Ryder, Subbiah, Kaplan, Dhariwal, Neelakantan, Shyam, Sastry, Askell, Agarwal, Herbert-Voss, Krueger, Henighan, Child, Ramesh, Ziegler, Wu, Winter, Hesse, Chen, Sigler, Litwin, Gray, Chess, Clark, Berner, McCandlish, Radford, Sutskever, and Amodei}]{llms_fewshot}
Tom Brown, Benjamin Mann, Nick Ryder, Melanie Subbiah, Jared~D Kaplan, Prafulla Dhariwal, Arvind Neelakantan, Pranav Shyam, Girish Sastry, Amanda Askell, Sandhini Agarwal, Ariel Herbert-Voss, Gretchen Krueger, Tom Henighan, Rewon Child, Aditya Ramesh, Daniel Ziegler, Jeffrey Wu, Clemens Winter, and 12 others. 2020.
\newblock \href {https://proceedings.neurips.cc/paper_files/paper/2020/file/1457c0d6bfcb4967418bfb8ac142f64a-Paper.pdf} {Language models are few-shot learners}.
\newblock In \emph{Advances in Neural Information Processing Systems}, volume~33, pages 1877--1901. Curran Associates, Inc.

\bibitem[{Cao et~al.(2023)Cao, Hao, Chen, Liu, Xu, Li, Jiang, and Zhao}]{cao-etal-2023-event}
Pengfei Cao, Yupu Hao, Yubo Chen, Kang Liu, Jiexin Xu, Huaijun Li, Xiaojian Jiang, and Jun Zhao. 2023.
\newblock \href {https://doi.org/10.18653/v1/2023.emnlp-main.21} {Event ontology completion with hierarchical structure evolution networks}.
\newblock In \emph{Proceedings of the 2023 Conference on Empirical Methods in Natural Language Processing}, pages 306--320, Singapore. Association for Computational Linguistics.

\bibitem[{Casati and Varzi(2025)}]{sep-events}
Roberto Casati and Achille Varzi. 2025.
\newblock {Events}.
\newblock In Edward~N. Zalta and Uri Nodelman, editors, \emph{The {Stanford} Encyclopedia of Philosophy}, {S}ummer 2025 edition. Metaphysics Research Lab, Stanford University.

\bibitem[{Chambers and Jurafsky(2011)}]{chambers-jurafsky-2011-template}
Nathanael Chambers and Dan Jurafsky. 2011.
\newblock \href {https://aclanthology.org/P11-1098/} {Template-based information extraction without the templates}.
\newblock In \emph{Proceedings of the 49th Annual Meeting of the Association for Computational Linguistics: Human Language Technologies}, pages 976--986, Portland, Oregon, USA. Association for Computational Linguistics.

\bibitem[{Chen et~al.(2024{\natexlab{a}})Chen, Xiao, Zhang, Luo, Lian, and Liu}]{chen-etal-2024-m3}
Jianlyu Chen, Shitao Xiao, Peitian Zhang, Kun Luo, Defu Lian, and Zheng Liu. 2024{\natexlab{a}}.
\newblock \href {https://doi.org/10.18653/v1/2024.findings-acl.137} {{M}3-embedding: Multi-linguality, multi-functionality, multi-granularity text embeddings through self-knowledge distillation}.
\newblock In \emph{Findings of the Association for Computational Linguistics: ACL 2024}, pages 2318--2335, Bangkok, Thailand. Association for Computational Linguistics.

\bibitem[{Chen et~al.(2024{\natexlab{b}})Chen, Qin, Jiang, and Choi}]{Chen_Qin_Jiang_Choi_2024}
Ruirui Chen, Chengwei Qin, Weifeng Jiang, and Dongkyu Choi. 2024{\natexlab{b}}.
\newblock \href {https://doi.org/10.1609/aaai.v38i16.29730} {Is a large language model a good annotator for event extraction?}
\newblock \emph{Proceedings of the AAAI Conference on Artificial Intelligence}, 38(16):17772--17780.

\bibitem[{Devlin et~al.(2019)Devlin, Chang, Lee, and Toutanova}]{devlin-etal-2019-bert}
Jacob Devlin, Ming-Wei Chang, Kenton Lee, and Kristina Toutanova. 2019.
\newblock \href {https://doi.org/10.18653/v1/N19-1423} {{BERT}: Pre-training of deep bidirectional transformers for language understanding}.
\newblock In \emph{Proceedings of the 2019 Conference of the North {A}merican Chapter of the Association for Computational Linguistics: Human Language Technologies, Volume 1 (Long and Short Papers)}, pages 4171--4186, Minneapolis, Minnesota. Association for Computational Linguistics.

\bibitem[{Doddington et~al.(2004)Doddington, Mitchell, Przybocki, Ramshaw, Strassel, and Weischedel}]{doddington-etal-2004-automatic}
George Doddington, Alexis Mitchell, Mark Przybocki, Lance Ramshaw, Stephanie Strassel, and Ralph Weischedel. 2004.
\newblock \href {https://aclanthology.org/L04-1011/} {The automatic content extraction ({ACE}) program {--} tasks, data, and evaluation}.
\newblock In \emph{Proceedings of the Fourth International Conference on Language Resources and Evaluation ({LREC}{'}04)}, Lisbon, Portugal. European Language Resources Association (ELRA).

\bibitem[{Dong et~al.(2025)Dong, Shah, Sonawane, Banerjee, Brady, Du, and Jiang}]{dong-etal-2025-scievent}
Bofu Dong, Pritesh Shah, Sumedh Sonawane, Tiyasha Banerjee, Erin Brady, Xinya Du, and Ming Jiang. 2025.
\newblock \href {https://doi.org/10.18653/v1/2025.emnlp-main.871} {{S}ci{E}vent: Benchmarking multi-domain scientific event extraction}.
\newblock In \emph{Proceedings of the 2025 Conference on Empirical Methods in Natural Language Processing}, pages 17225--17255, Suzhou, China. Association for Computational Linguistics.

\bibitem[{Ebner et~al.(2020)Ebner, Xia, Culkin, Rawlins, and Van~Durme}]{ebner-etal-2020-multi}
Seth Ebner, Patrick Xia, Ryan Culkin, Kyle Rawlins, and Benjamin Van~Durme. 2020.
\newblock \href {https://doi.org/10.18653/v1/2020.acl-main.718} {Multi-sentence argument linking}.
\newblock In \emph{Proceedings of the 58th Annual Meeting of the Association for Computational Linguistics}, pages 8057--8077, Online. Association for Computational Linguistics.

\bibitem[{Fedoryszak et~al.(2019)Fedoryszak, Frederick, Rajaram, and Zhong}]{rt_ed}
Mateusz Fedoryszak, Brent Frederick, Vijay Rajaram, and Changtao Zhong. 2019.
\newblock \href {https://doi.org/10.1145/3292500.3330689} {Real-time event detection on social data streams}.
\newblock In \emph{Proceedings of the 25th ACM SIGKDD International Conference on Knowledge Discovery \& Data Mining (KDD '19)}, pages 2774--2782, New York, NY, USA. Association for Computing Machinery.

\bibitem[{Guo et~al.(2023)Guo, Wang, Chen, Liu, Li, and Zhao}]{guo-etal-2023-eventoa}
Shaoru Guo, Chenhao Wang, Yubo Chen, Kang Liu, Ru~Li, and Jun Zhao. 2023.
\newblock \href {https://doi.org/10.18653/v1/2023.findings-acl.637} {{E}vent{OA}: An event ontology alignment benchmark based on {F}rame{N}et and {W}ikidata}.
\newblock In \emph{Findings of the Association for Computational Linguistics: ACL 2023}, pages 10038--10052, Toronto, Canada. Association for Computational Linguistics.

\bibitem[{He et~al.(2019)He, Fan, Wu, Xie, and Girshick}]{DBLP:journals/corr/abs-1911-05722}
Kaiming He, Haoqi Fan, Yuxin Wu, Saining Xie, and Ross~B. Girshick. 2019.
\newblock \href {https://arxiv.org/abs/1911.05722} {Momentum contrast for unsupervised visual representation learning}.
\newblock \emph{CoRR}, abs/1911.05722.

\bibitem[{Hsu et~al.(2022)Hsu, Huang, Boschee, Miller, Natarajan, Chang, and Peng}]{hsu-etal-2022-degree}
I-Hung Hsu, Kuan-Hao Huang, Elizabeth Boschee, Scott Miller, Prem Natarajan, Kai-Wei Chang, and Nanyun Peng. 2022.
\newblock \href {https://doi.org/10.18653/v1/2022.naacl-main.138} {{DEGREE}: A data-efficient generation-based event extraction model}.
\newblock In \emph{Proceedings of the 2022 Conference of the North American Chapter of the Association for Computational Linguistics: Human Language Technologies}, pages 1890--1908, Seattle, United States. Association for Computational Linguistics.

\bibitem[{Huang et~al.(2024)Huang, Hsu, Parekh, Xie, Zhang, Natarajan, Chang, Peng, and Ji}]{huang-etal-2024-textee}
Kuan-Hao Huang, I-Hung Hsu, Tanmay Parekh, Zhiyu Xie, Zixuan Zhang, Prem Natarajan, Kai-Wei Chang, Nanyun Peng, and Heng Ji. 2024.
\newblock \href {https://doi.org/10.18653/v1/2024.findings-acl.760} {{T}ext{EE}: Benchmark, reevaluation, reflections, and future challenges in event extraction}.
\newblock In \emph{Findings of the Association for Computational Linguistics: ACL 2024}, pages 12804--12825, Bangkok, Thailand. Association for Computational Linguistics.

\bibitem[{Huang and Ji(2020)}]{huang-ji-2020-semi}
Lifu Huang and Heng Ji. 2020.
\newblock \href {https://doi.org/10.18653/v1/2020.emnlp-main.53} {Semi-supervised new event type induction and event detection}.
\newblock In \emph{Proceedings of the 2020 Conference on Empirical Methods in Natural Language Processing (EMNLP)}, pages 718--724, Online. Association for Computational Linguistics.

\bibitem[{Hubert and Arabie(1985)}]{ari}
Lawrence Hubert and Phipps Arabie. 1985.
\newblock \href {https://doi.org/10.1007/BF01908075} {Comparing partitions}.
\newblock \emph{Journal of Classification}, 2(1):193--218.

\bibitem[{Kim et~al.(2013)Kim, Wang, and Yasunori}]{kim-etal-2013-genia}
Jin-Dong Kim, Yue Wang, and Yamamoto Yasunori. 2013.
\newblock \href {https://aclanthology.org/W13-2002/} {The {G}enia event extraction shared task, 2013 edition - overview}.
\newblock In \emph{Proceedings of the {B}io{NLP} Shared Task 2013 Workshop}, pages 8--15, Sofia, Bulgaria. Association for Computational Linguistics.

\bibitem[{Li et~al.(2021{\natexlab{a}})Li, Zhou, Xiong, and Hoi}]{DBLP:conf/iclr/0001ZXH21}
Junnan Li, Pan Zhou, Caiming Xiong, and Steven C.~H. Hoi. 2021{\natexlab{a}}.
\newblock \href {https://openreview.net/forum?id=KmykpuSrjcq} {Prototypical contrastive learning of unsupervised representations}.
\newblock In \emph{9th International Conference on Learning Representations, {ICLR} 2021, Virtual Event, Austria, May 3-7, 2021}. OpenReview.net.

\bibitem[{Li et~al.(2020)Li, Zareian, Zeng, Whitehead, Lu, Ji, and Chang}]{li-etal-2020-cross}
Manling Li, Alireza Zareian, Qi~Zeng, Spencer Whitehead, Di~Lu, Heng Ji, and Shih-Fu Chang. 2020.
\newblock \href {https://doi.org/10.18653/v1/2020.acl-main.230} {Cross-media structured common space for multimedia event extraction}.
\newblock In \emph{Proceedings of the 58th Annual Meeting of the Association for Computational Linguistics}, pages 2557--2568, Online. Association for Computational Linguistics.

\bibitem[{Li et~al.(2021{\natexlab{b}})Li, Ji, and Han}]{li-etal-2021-document}
Sha Li, Heng Ji, and Jiawei Han. 2021{\natexlab{b}}.
\newblock \href {https://doi.org/10.18653/v1/2021.naacl-main.69} {Document-level event argument extraction by conditional generation}.
\newblock In \emph{Proceedings of the 2021 Conference of the North American Chapter of the Association for Computational Linguistics: Human Language Technologies}, pages 894--908, Online. Association for Computational Linguistics.

\bibitem[{Li et~al.(2022)Li, Ji, and Han}]{li-etal-2022-open}
Sha Li, Heng Ji, and Jiawei Han. 2022.
\newblock \href {https://doi.org/10.18653/v1/2022.emnlp-main.461} {Open relation and event type discovery with type abstraction}.
\newblock In \emph{Proceedings of the 2022 Conference on Empirical Methods in Natural Language Processing}, pages 6864--6877, Abu Dhabi, United Arab Emirates. Association for Computational Linguistics.

\bibitem[{Li et~al.(2023)Li, Zhan, Conger, Palmer, Ji, and Han}]{li-etal-2023-glen}
Sha Li, Qiusi Zhan, Kathryn Conger, Martha Palmer, Heng Ji, and Jiawei Han. 2023.
\newblock \href {https://doi.org/10.18653/v1/2023.emnlp-main.170} {{GLEN}: General-purpose event detection for thousands of types}.
\newblock In \emph{Proceedings of the 2023 Conference on Empirical Methods in Natural Language Processing}, pages 2823--2838, Singapore. Association for Computational Linguistics.

\bibitem[{Lin(2004)}]{lin-2004-rouge}
Chin-Yew Lin. 2004.
\newblock \href {https://aclanthology.org/W04-1013/} {{ROUGE}: A package for automatic evaluation of summaries}.
\newblock In \emph{Text Summarization Branches Out}, pages 74--81, Barcelona, Spain. Association for Computational Linguistics.

\bibitem[{Lin et~al.(2020)Lin, Ji, Huang, and Wu}]{lin-etal-2020-joint}
Ying Lin, Heng Ji, Fei Huang, and Lingfei Wu. 2020.
\newblock \href {https://doi.org/10.18653/v1/2020.acl-main.713} {A joint neural model for information extraction with global features}.
\newblock In \emph{Proceedings of the 58th Annual Meeting of the Association for Computational Linguistics}, pages 7999--8009, Online. Association for Computational Linguistics.

\bibitem[{Lu et~al.(2023)Lu, Ran, Tetreault, and Jaimes}]{lu-etal-2023-event}
Di~Lu, Shihao Ran, Joel Tetreault, and Alejandro Jaimes. 2023.
\newblock \href {https://doi.org/10.18653/v1/2023.acl-short.143} {Event extraction as question generation and answering}.
\newblock In \emph{Proceedings of the 61st Annual Meeting of the Association for Computational Linguistics (Volume 2: Short Papers)}, pages 1666--1688, Toronto, Canada. Association for Computational Linguistics.

\bibitem[{Lu et~al.(2025)Lu, Mao, Lan, Zhang, Zhu, and Huang}]{lu-etal-2025-seoe}
Yi-Fan Lu, Xian-Ling Mao, Tian Lan, Tong Zhang, Yu-Shi Zhu, and Heyan Huang. 2025.
\newblock \href {https://doi.org/10.18653/v1/2025.acl-long.356} {{SEOE}: A scalable and reliable semantic evaluation framework for open domain event detection}.
\newblock In \emph{Proceedings of the 63rd Annual Meeting of the Association for Computational Linguistics (Volume 1: Long Papers)}, pages 7201--7218, Vienna, Austria. Association for Computational Linguistics.

\bibitem[{Nguyen et~al.(2015)Nguyen, Tannier, Ferret, and Besan{\c{c}}on}]{nguyen-etal-2015-generative}
Kiem-Hieu Nguyen, Xavier Tannier, Olivier Ferret, and Romaric Besan{\c{c}}on. 2015.
\newblock \href {https://doi.org/10.3115/v1/P15-1019} {Generative event schema induction with entity disambiguation}.
\newblock In \emph{Proceedings of the 53rd Annual Meeting of the Association for Computational Linguistics and the 7th International Joint Conference on Natural Language Processing (Volume 1: Long Papers)}, pages 188--197, Beijing, China. Association for Computational Linguistics.

\bibitem[{Parekh et~al.(2025)Parekh, Dong, Bandarkar, Kim, Hsu, Chang, and Peng}]{parekh-etal-2025-snare}
Tanmay Parekh, Yuxuan Dong, Lucas Bandarkar, Artin Kim, I-Hung Hsu, Kai-Wei Chang, and Nanyun Peng. 2025.
\newblock \href {https://doi.org/10.18653/v1/2025.emnlp-main.1039} {{SN}a{R}e: Domain-aware data generation for low-resource event detection}.
\newblock In \emph{Proceedings of the 2025 Conference on Empirical Methods in Natural Language Processing}, pages 20583--20604, Suzhou, China. Association for Computational Linguistics.

\bibitem[{Peng et~al.(2023)Peng, Wang, Yao, Zeng, Hou, Li, Liu, and Shen}]{peng-etal-2023-devil}
Hao Peng, Xiaozhi Wang, Feng Yao, Kaisheng Zeng, Lei Hou, Juanzi Li, Zhiyuan Liu, and Weixing Shen. 2023.
\newblock \href {https://doi.org/10.18653/v1/2023.findings-acl.586} {The devil is in the details: On the pitfalls of event extraction evaluation}.
\newblock In \emph{Findings of the Association for Computational Linguistics: ACL 2023}, pages 9206--9227, Toronto, Canada. Association for Computational Linguistics.

\bibitem[{Reimers and Gurevych(2019)}]{reimers-gurevych-2019-sentence}
Nils Reimers and Iryna Gurevych. 2019.
\newblock \href {https://doi.org/10.18653/v1/D19-1410} {Sentence-{BERT}: Sentence embeddings using {S}iamese {BERT}-networks}.
\newblock In \emph{Proceedings of the 2019 Conference on Empirical Methods in Natural Language Processing and the 9th International Joint Conference on Natural Language Processing (EMNLP-IJCNLP)}, pages 3982--3992, Hong Kong, China. Association for Computational Linguistics.

\bibitem[{Satyapanich et~al.(2020)Satyapanich, Ferraro, and Finin}]{casie}
Taneeya Satyapanich, Francis Ferraro, and Tim Finin. 2020.
\newblock \href {https://doi.org/10.1609/aaai.v34i05.6401} {Casie: Extracting cybersecurity event information from text}.
\newblock \emph{Proceedings of the AAAI Conference on Artificial Intelligence}, 34(05):8749--8757.

\bibitem[{Segers et~al.(2017)Segers, Caselli, and Vossen}]{segers-etal-2017-circumstantial}
Roxane Segers, Tommaso Caselli, and Piek Vossen. 2017.
\newblock \href {https://doi.org/10.18653/v1/W17-2706} {The circumstantial event ontology ({CEO})}.
\newblock In \emph{Proceedings of the Events and Stories in the News Workshop}, pages 37--41, Vancouver, Canada. Association for Computational Linguistics.

\bibitem[{Shen et~al.(2021)Shen, Zhang, Ji, and Han}]{shen-etal-2021-corpus}
Jiaming Shen, Yunyi Zhang, Heng Ji, and Jiawei Han. 2021.
\newblock \href {https://doi.org/10.18653/v1/2021.emnlp-main.441} {Corpus-based open-domain event type induction}.
\newblock In \emph{Proceedings of the 2021 Conference on Empirical Methods in Natural Language Processing}, pages 5427--5440, Online and Punta Cana, Dominican Republic. Association for Computational Linguistics.

\bibitem[{Sun et~al.(2022)Sun, Li, Pergola, Wallace, John, Greene, Kim, and He}]{sun-etal-2022-phee}
Zhaoyue Sun, Jiazheng Li, Gabriele Pergola, Byron Wallace, Bino John, Nigel Greene, Joseph Kim, and Yulan He. 2022.
\newblock \href {https://doi.org/10.18653/v1/2022.emnlp-main.376} {{PHEE}: A dataset for pharmacovigilance event extraction from text}.
\newblock In \emph{Proceedings of the 2022 Conference on Empirical Methods in Natural Language Processing}, pages 5571--5587, Abu Dhabi, United Arab Emirates. Association for Computational Linguistics.

\bibitem[{Sundheim(1992)}]{sundheim-1992-overview}
Beth~M. Sundheim. 1992.
\newblock \href {https://aclanthology.org/M92-1001/} {Overview of the fourth {M}essage {U}nderstanding {E}valuation and {C}onference}.
\newblock In \emph{{F}ourth {M}essage {U}nderstanding {C}onference ({MUC}-4): Proceedings of a Conference Held in {M}c{L}ean, {V}irginia, {J}une 16-18, 1992}.

\bibitem[{van~den Oord et~al.(2018)van~den Oord, Li, and Vinyals}]{Oord2018RepresentationLW}
A{\"a}ron van~den Oord, Yazhe Li, and Oriol Vinyals. 2018.
\newblock Representation learning with contrastive predictive coding.
\newblock \emph{ArXiv}, abs/1807.03748.

\bibitem[{Wadden et~al.(2019)Wadden, Wennberg, Luan, and Hajishirzi}]{wadden-etal-2019-entity}
David Wadden, Ulme Wennberg, Yi~Luan, and Hannaneh Hajishirzi. 2019.
\newblock \href {https://doi.org/10.18653/v1/D19-1585} {Entity, relation, and event extraction with contextualized span representations}.
\newblock In \emph{Proceedings of the 2019 Conference on Empirical Methods in Natural Language Processing and the 9th International Joint Conference on Natural Language Processing (EMNLP-IJCNLP)}, pages 5784--5789, Hong Kong, China. Association for Computational Linguistics.

\bibitem[{Wang et~al.(2023)Wang, Zhou, Qiao, Li, and Li}]{wang-etal-2023-improving-unsupervised}
Qing Wang, Kang Zhou, Qiao Qiao, Yuepei Li, and Qi~Li. 2023.
\newblock \href {https://doi.org/10.18653/v1/2023.emnlp-main.745} {Improving unsupervised relation extraction by augmenting diverse sentence pairs}.
\newblock In \emph{Proceedings of the 2023 Conference on Empirical Methods in Natural Language Processing}, pages 12136--12147, Singapore. Association for Computational Linguistics.

\bibitem[{Wang et~al.(2020)Wang, Wang, Han, Jiang, Han, Liu, Li, Li, Lin, and Zhou}]{wang-etal-2020-maven}
Xiaozhi Wang, Ziqi Wang, Xu~Han, Wangyi Jiang, Rong Han, Zhiyuan Liu, Juanzi Li, Peng Li, Yankai Lin, and Jie Zhou. 2020.
\newblock \href {https://doi.org/10.18653/v1/2020.emnlp-main.129} {{MAVEN}: {A} {M}assive {G}eneral {D}omain {E}vent {D}etection {D}ataset}.
\newblock In \emph{Proceedings of the 2020 Conference on Empirical Methods in Natural Language Processing (EMNLP)}, pages 1652--1671, Online. Association for Computational Linguistics.

\bibitem[{Wei et~al.(2025)Wei, Shi, Tong, Reddy, Natarajan, Jain, Garimella, and Huang}]{wei-etal-2025-legalcore}
Kangda Wei, Xi~Shi, Jonathan Tong, Sai~Ramana Reddy, Anandhavelu Natarajan, Rajiv Jain, Aparna Garimella, and Ruihong Huang. 2025.
\newblock \href {https://doi.org/10.18653/v1/2025.findings-acl.1284} {{L}egal{C}ore: A dataset for event coreference resolution in legal documents}.
\newblock In \emph{Findings of the Association for Computational Linguistics: ACL 2025}, pages 25044--25059, Vienna, Austria. Association for Computational Linguistics.

\bibitem[{Xu et~al.(2024)Xu, Zhang, and Chen}]{xu-etal-2024-ceo}
Nan Xu, Hongming Zhang, and Jianshu Chen. 2024.
\newblock \href {https://aclanthology.org/2024.findings-eacl.64/} {{CEO}: Corpus-based open-domain event ontology induction}.
\newblock In \emph{Findings of the Association for Computational Linguistics: EACL 2024}, pages 946--964, St. Julian{'}s, Malta. Association for Computational Linguistics.

\bibitem[{Yuan et~al.(2018)Yuan, Ren, He, Zhang, Geng, Huang, Ji, Lin, and Han}]{news_event}
Quan Yuan, Xiang Ren, Wenqi He, Chao Zhang, Xinhe Geng, Lifu Huang, Heng Ji, Chin-Yew Lin, and Jiawei Han. 2018.
\newblock \href {https://doi.org/10.1145/3269206.3271674} {Open-schema event profiling for massive news corpora}.
\newblock In \emph{Proceedings of the 27th ACM International Conference on Information and Knowledge Management}, CIKM '18, page 587–596, New York, NY, USA. Association for Computing Machinery.

\bibitem[{Zhang* et~al.(2020)Zhang*, Kishore*, Wu*, Weinberger, and Artzi}]{bert-score}
Tianyi Zhang*, Varsha Kishore*, Felix Wu*, Kilian~Q. Weinberger, and Yoav Artzi. 2020.
\newblock \href {https://openreview.net/forum?id=SkeHuCVFDr} {Bertscore: Evaluating text generation with bert}.
\newblock In \emph{International Conference on Learning Representations}.

\end{thebibliography}


\clearpage
\end{document}